# Banana Sub-Family Classification and Quality Prediction using Computer Vision

Narayana Darapaneni[1], Arjun Tanndalam[2], Mohit Gupta[3], Neeta Taneja[4], Prabu Purushothaman[5], Swati Eswar[6], Anwesh Reddy Paduri[7], Thangaselvi Arichandrapandian[8]

[1]*Northwestern University/Great Learning, US*

[2-8] *Great Learning, Bangalore, India*

anwesh@greatlearning.in

*Abstract:* India is the second largest producer of fruits and vegetables in the world, and one of the largest consumers of fruits like Banana, Papaya and Mangoes through retail and ecommerce giants like BigBasket, Grofers and Amazon Fresh. However, adoption of technology in supply chain and retail stores is still low and there is a great potential to adopt computer-vision based technology for identification and classification of fruits. We have chosen banana fruit to build a computer vision based model to carry out the following three use-cases (a) Identify Banana from a given image (b) Determine sub-family or variety of Banana (c) Determine the quality of Banana. Successful execution of these use-cases using computer-vision model would greatly help with overall inventory management automation, quality control, quick and efficient weighing and billing which all are manual labor intensive currently. In this work, we suggest a machine learning pipeline that combines the ideas of CNNs, transfer learning, and data augmentation towards improving Banana fruit sub family and quality image classification. We have built a basic CNN and then went on to tune a MobileNet Banana classification model using a combination of self-curated and publicly-available dataset of 3064 images. The results show an overall 93.4% and 100% accuracy for sub-family/variety and for quality test classifications respectively.

*Keywords:* Banana; Variety; Quality; Computer Vision; Data Augmentation; Classification

## 1. Introduction

Recognition of fruit type and quality and its automation is important in smart agriculture to increase production efficiency. As of 2018-19, total production stood at 313 million tons. Among fruits, India ranks first in the production of fruits like bananas (25.7%), papayas (43.6%) and mangoes (40.4%). This high production serves both domestic consumption, high amounts of exports and also processed food. Domestic purchase and consumption are being slowly penetrated by eCommerce delivery giants like BigBasket, Grofers and Amazon Fresh who rely on technology not just for the commerce platform but also in entire supply chain, packaging and delivery. At the same time, retailer shops are also gradually adopting technology at quality checks, weighing and point of sale. Less manual interventions at various stages both to save cost and reduce manual handling are gaining importance, more so after COVID-induced behaviors became the norm.

One of the important use-case in this technology adoption for both eCommerce and Retailer scenario we are exploring is identification of fruit using computer vision, ability to decipher the sub-family (or variety) of the fruit and predict the quality of the fruit. For the purpose of this work of research the fruit we picked is Banana given its popularity in consumption and number of varieties available. The three main use cases we have incorporated in this work are:

- Identify Banana from a given image
- Determine sub-family or variety of Banana
- Determine the quality of Banana

Currently this is done completely manually by most of the large eCommerce players and at Retailer shops. We intend to solve this problem using Deep-Learning techniques available. Successful development of the solution can be leveraged to integrate with existing ecosystem of products and methods in use currently. The idea can be extended to multiple fruits, vegetables and other inventory products and multiple products simultaneously. It can be a primary step for overall inventory management automation, quality control, quick and efficient weighing and billing which all are manual labor intensive currently. The major scientific contributions we present in this paper are as follows:

- Exploration of Convolutional Neural Network (CNN) topologies for fruit classification.
- Data augmentation of manually collected images and existing publicly-available datasets.
- Implementation of a Convolutional Neural Network (CNN) model for the classification of obtained Banana images.
- Implementation of transfer learning using MobileNet, VGGNet, EfficientNet, ResNet & InceptionNet.
- Converged to choose MobileNet architecture
- Further model tuning shows that a MobileNet classification network can retain 93.4% classification accuracy.

## 2. Literature Survey

To predict the quality of fruit or vegetable using images there are several approaches published. Learning from the references mentioned in this section. The key takeaway from the existing work was the model architectures and hyper parameters. In [6], published in 2014 Zhang and team have used a variant of ANNs to recognize fruits. Fruits images were acquired by a digital camera and then the background of each image was removed by split-and-merge algorithm and further resized/down sampled to 256x256. The color histogram, texture and shape features of each fruit image were extracted to compose a feature space. In total, there are 79 features (64 color features + 7 texture features + 8 shape features) extracted from a prescribed image. PCA is applied on this feature space to reduce dimensionality and then it is split into training and test sets. The experimental results on the 1653 color fruit images from the 18 categories of fruits achieved a classification accuracy of 89.1%. *Fruit recognition from images using deep learning 2 Dec 2017 · Horea Mureşan, Mihai Oltean* [18] Introduces the Fruit 360 dataset with high quality images of different fruits. The dataset contains 61934 images of fruits spanning across 90 fruits. The fruits are planted on a low speed motor with a speed of 3rpm and a short video is recorded for 20 secs which is then broken into different frames and added as a dataset. A white sheet of paper is used as the background for collecting the images. A separate algorithm was applied to get the background appear the same in different lighting conditions.

*Automatic Fruit Classification Using Deep Learning for Industrial Applications 10 October 2018, M. Shamim Hossain; Muneer Al-Hammadi; Ghulam Muhammad.* [19] In this paper, authors built a framework for fruit classification using deep learning. Framework is based on two different deep learning architectures. The first is a proposed light model of six convolutional neural network layers, whereas the second is a fine-tuned visual geometry group-16 pre trained deep learning model. Two color image datasets, one of which is publicly available, are used to evaluate the proposed framework. The first dataset (dataset 1) consists of clear fruit images, whereas the second dataset (dataset 2) contains fruit images that are challenging to classify. Classification accuracies of 99.49% and 99.75% were achieved on dataset 1 for the first and second models, respectively. On dataset 2, the first and second models obtained accuracies of 85.43% and 96.75%, respectively. *A Hierarchical Grocery Store Image Dataset with Visual and Semantic Labels 03 Jan 2019, Marcus Klasson, Cheng Zhang, Hedvig Kjellström* - [3] In this paper authors have tried to implement image classification models and provide a benchmark dataset for classification of fruits and vegetables and other grocery items like packets of curd and milk. The idea was to help visually impaired people to shop in a grocery store. Pre trained convolutional network is used along with multi view variational autoencoder, which has captured information available in the dataset. There are a total of 5125 natural images used in the paper with 81 classes. The classes are further subdivided into multiple levels, coarse grained classification into classes like apple , orange etc. Further the fine grained classes like type of apple e.g. fuji, red apple etc. The best accuracy was achieved using DenseNet-169 of around 85.2% got coarse grained classes. *Fruit Quality and Defect Image Classification with Conditional GAN Data Augmentation 12 Apr 2021 by Jordan J. Bird, Chloe M. Barnes, Luis J. Manso, Anikó Ekárt, Diego R. Faria* [5] have tried to judge the quality of lemons using the images from Fruit 360 dataset using a publicly-available dataset of 2690 images. Due to lack of data they have used GANs to enrich their dataset. It proposed the use of pretrained VGG16 as a model with a small batch size to accommodate it on commodity hardware. It was found that appending a 4096 neuron fully connected layer to the convolutional layers leads to an image classification accuracy of 83.77%. The model was then trained on a Conditional Generative Adversarial Network on the training data for 2000 epochs, and it learned to generate relatively realistic images. Grad-CAM analysis of the model trained on real photographs shows that the synthetic images can exhibit classifiable characteristics such as shape, mould, and gangrene. A higher image classification accuracy of 88.75% is then attained by augmenting the training with synthetic images.

*Ripeness Classification of Bananas Using an Artificial Neural Network January 2019, Ahmed Nashat, Fatma Mazen* [11] In this paper, an automatic computer vision system is proposed to identify the ripening stages of bananas. First, a four-class homemade database is prepared. Second, an artificial neural network-based framework which uses color, development of brown spots, and Tamura statistical texture features is employed to classify and grade banana fruit ripening stage. Results and the performance of the proposed system are compared with various techniques such as the SVM, the naive Bayes, the KNN, the decision tree, and discriminant analysis classifiers. Results reveal that the proposed system has the highest overall recognition rate, which is 97.75%, among other techniques. *Classification of Banana Fruits Using Deep Learning, 2020, Ahmed F. Al-daour, Mohammed O. Al-shawwa.* [20] In this paper, machine learning based approach is presented for identifying type of banana with a dataset that contains 8,554 images, out of which 4,488 images were used for training, 1,928 images for validation and 2,138 images for testing. A deep learning technique that extensively applied to image recognition was used. Using 70% from image for training and 30% from image for validation their trained model achieved an accuracy of 100% on a held-out test set, demonstrating the feasibility of this approach.

## 3. Method

### 3.1. Data collection and pre processing

We explored using existing dataset like Fruits 360 to use some existing available data. However, on assessment of existing open datasets, we realized that to make this useful for the Indian context and also to address the quality prediction use-case, we needed some new data which needs to be collected from scratch. We addressed this by procuring different sub-families of bananas from the market, clicked pictures of them and created our own dataset required for the project. We created about 1500+ images of the banana dataset using these pictures. The sub-families that were used are Elakki, Red Banana, Robusta, Nendram and Hill Banana. We further augmented this by using about 800 banana images data from Fruits 360 dataset for Robusta and Red Banana. We also added some negative use-case images of other fruits, so that we can address the first use-case of identifying Banana from the rest of the fruits. For this purpose, we clicked pictures of fruits like sweet lime, pomegranate and apple and added them as Other Fruits in the dataset. All the pictures were clicked with plain background so that the noise level in learning is reduced, and below Table 1 are the specifications of the camera used:

| Resolution | 4608 x 2592 |
|---|---|
| File Size | 2MB to 4MB |
| Focal Length | 4.1 mm |
| Aperture | f/1.7 |
| Exposure Time | 1/33 |
| ISO | 1000 |

Table 1: Specifications of camera used

*3.2. Exploratory Data Analysis*

Dataset created for the project has a total of 3064 images and 6 classes. The number of images for each of the class is available in the graph below. Some examples of the images from our dataset are also shown below in Figure 1 and Figure

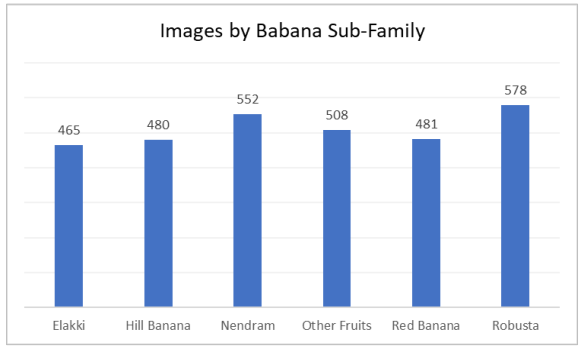

Figure 1 Dataset size of six varieties of fruits

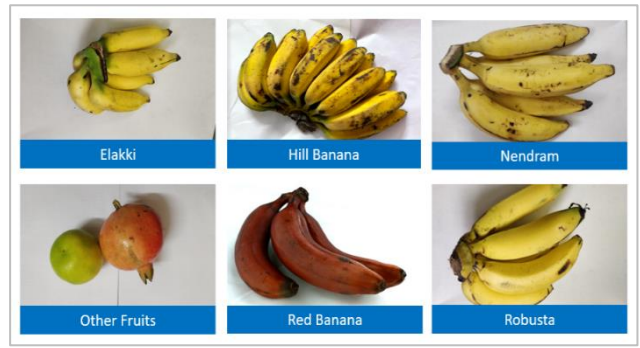

Figure 2 Sample images of each variety of fruit including additional sample of bad quality Elakki banana

*3.3. Data Pre-processing and Augmentation*

We split the data collected into train and test data folders for us to have some clean unseen data that we can use for testing. We have ensured a suitable naming convention that would assist us in labelling our data using respective folder names. We used Image Data Generator's flow from directory method to read data and label them for all three data sets of train, validation and test. Around 76% data was used for training, around 19% data was used for validation and approximately 5% data was used for testing purposes.

Image augmentation techniques are applied at random on images to add diversity and increase data volume. The operations performed in augmentation are rotation, shifting and flip (horizontally and vertically) before providing it to the model. Figure 1 are image samples of augmentation applied on the dataset:

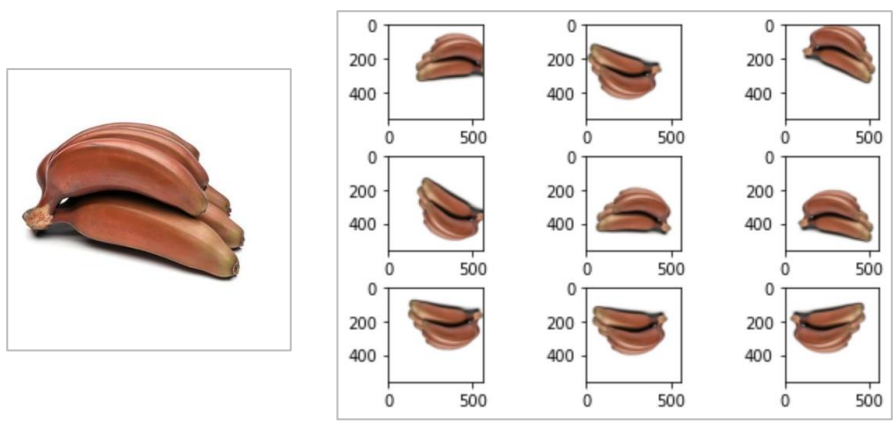

Figure 1 Depiction of original image (left) and augmented images (right) on a sample of Red Banana

*3.4. Model Building*

Our model development process has two stages – first we implemented a base CNN model with below mentioned architecture and second phase was implementing transfer learning using various available pre-trained models such as MobileNet, VGGNet, EfficientNet, ResNet & InceptionNet. The following sections describe further details on both stages of the process.

Additionally, we have made the following observations as part of the model building and tuning process
- Increasing the number of epochs improves the accuracy
- Early stopping helps to reduce the execution time
- Changes in batch size improves execution time and memory requirement
- Augmentation helps in avoiding overfitting of data

*3.4.1. Base CNN Model*

We have built a base CNN model using Tensorflow Keras library. Image size is kept as 256 x 256 pixels. We onvolutions. The structure of MobileNet is shown in Figure 2.

have 1 input layer, 4 number of convolution layers, 4 number of max-pooling, 3 number of batch-normalization layer, 1 drop out layer, 3 Dense layers and then finally 1 output layer with 6 classes using softmax for classification. We defined model checkpoints to store weights after each run and stored the output in a history file. On executing this for 10 epochs, we received a model validation accuracy of 75.1%

To further improve the accuracy, we tried a few transfer-learning models like MobileNet, VGGNet, EfficientNet, ResNet & InceptionNet and the performance of which are shown in the results section of this paper.

*3.4.2. Transfer Learning*

*3.4.3.* Transfer learning is a machine learning method where a model developed for a task is reused as the starting point for a model on a second task. It is a popular approach in deep learning where pre-trained models are used as the starting point on computer vision and natural language processing tasks.

So, transfer learning was the go to strategy for improving model performance. The number of available architectures to train goes beyond count. Comparing all architectures to each other is a difficult task. After having tried the following MobileNet, VGGNet, EfficientNet, ResNet & InceptionNet, we were able to conclude that Inception Net, VGG Net and MobileNet architectures have properties making them worthwhile evaluating for this project.

*3.4.4. MobileNet Model*

MobileNet is a type of Convolution Neural Network Architecture used for Image Classification. It is a mobile-first model, which helps with higher accuracy consuming lower computation resources. MobileNet Architectures not only reduce the model size but improve prediction speed 10x with comparable accuracy. They use regular Convolutional layer but only once (at the beginning), all other layers use Depth-wise Separable Convolution. Depth-wise Separable Convolution is a combination of two Conv Layers - Depth-wise Convolution & Pointwise Convolution. The depthwise convolution filter performs a single convolution on each input channel, and the point convolution filter combines the output of depthwise convolution linearly with $1 * 1\ c$ convolutions. The structure of MobileNet is shown in Figure 2.

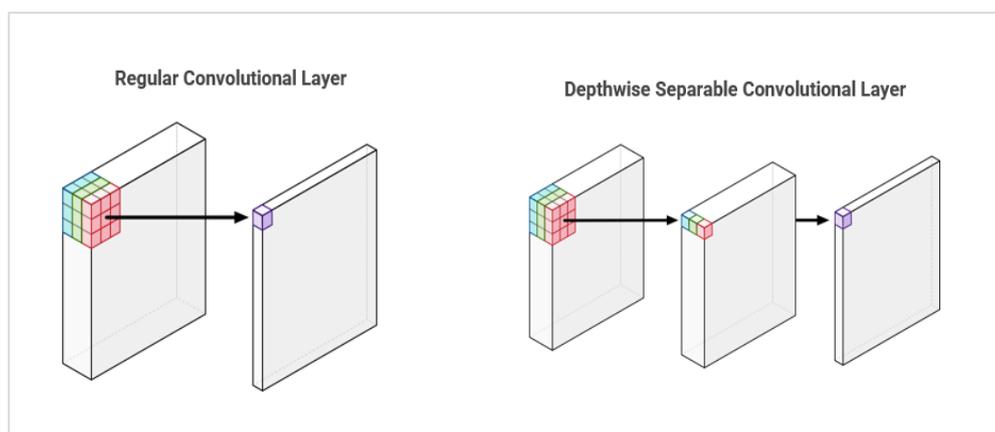

Figure 2 Comparing regular convolution layer with MobileNet convolution mechanism

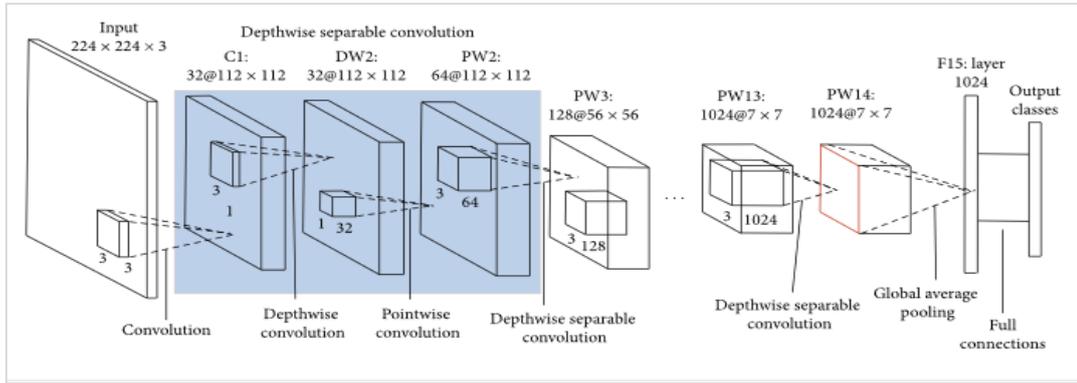

Figure 3 Image depicting depth wise separable convolution in MobileNet architecture

### 3.4.4 Model Architecture

In this work of research, we trained MobileNet with transfer learning using weights from a model trained with the ImageNet weights. For our model we have used pre-trained MobileNet model and replaced the last layer with three additional dense layers with 1024, 512 and 256 neurons, followed by an output layer for classification into the 6 classes of the banana variety in our dataset. The weights of the model built is also used for another of our use case of identifying and classification of the quality of banana for one of the varieties.

We set the first 20 layers as non-trainable and from 21$^{st}$ and layer up to the last layer as trainable. The architecture is depicted in the Figure 4

| Table 1. MobileNet Body Architecture | | |
|---|---|---|
| Type / Stride | Filter Shape | Input Size |
| Conv / s2 | $3 \times 3 \times 3 \times 32$ | $224 \times 224 \times 3$ |
| Conv dw / s1 | $3 \times 3 \times 32$ dw | $112 \times 112 \times 32$ |
| Conv / s1 | $1 \times 1 \times 32 \times 64$ | $112 \times 112 \times 32$ |
| Conv dw / s2 | $3 \times 3 \times 64$ dw | $112 \times 112 \times 64$ |
| Conv / s1 | $1 \times 1 \times 64 \times 128$ | $56 \times 56 \times 64$ |
| Conv dw / s1 | $3 \times 3 \times 128$ dw | $56 \times 56 \times 128$ |
| Conv / s1 | $1 \times 1 \times 128 \times 128$ | $56 \times 56 \times 128$ |
| Conv dw / s2 | $3 \times 3 \times 128$ dw | $56 \times 56 \times 128$ |
| Conv / s1 | $1 \times 1 \times 128 \times 256$ | $28 \times 28 \times 128$ |
| Conv dw / s1 | $3 \times 3 \times 256$ dw | $28 \times 28 \times 256$ |
| Conv / s1 | $1 \times 1 \times 256 \times 256$ | $28 \times 28 \times 256$ |
| Conv dw / s2 | $3 \times 3 \times 256$ dw | $28 \times 28 \times 256$ |
| Conv / s1 | $1 \times 1 \times 256 \times 512$ | $14 \times 14 \times 256$ |
| 5× Conv dw / s1 | $3 \times 3 \times 512$ dw | $14 \times 14 \times 512$ |
| Conv / s1 | $1 \times 1 \times 512 \times 512$ | $14 \times 14 \times 512$ |
| Conv dw / s2 | $3 \times 3 \times 512$ dw | $14 \times 14 \times 512$ |
| Conv / s1 | $1 \times 1 \times 512 \times 1024$ | $7 \times 7 \times 512$ |
| Conv dw / s2 | $3 \times 3 \times 1024$ dw | $7 \times 7 \times 1024$ |
| Conv / s1 | $1 \times 1 \times 1024 \times 1024$ | $7 \times 7 \times 1024$ |
| Avg Pool / s1 | Pool $7 \times 7$ | $7 \times 7 \times 1024$ |
| FC / s1 | $1024 \times 1000$ | $1 \times 1 \times 1024$ |
| Softmax / s1 | Classifier | $1 \times 1 \times 1000$ |

Figure 4 Model Architecture

The network was trained for 16 epochs. The models are trained using TensorFlow [1], with the implementation of MobileNet provided by Keras. The standard Adam Optimizer is used, with learning rate of 1e$^{-3}$. Table 2 lists the model attributes.

| Epochs | 16 |
|---|---|
| Optimizer | Adam |
| Loss | Categorical CrossEntrophy |

| Weights | imagenet |
|---|---|
| Global Average Pooling 2D Layer | |
| Dense Layer with 1024 Neurons | |
| Dense Layer with 512 Neurons | |
| Dense Layer with 256 Neurons | |
| Activation function Layer with SoftMax | |

Table 2 Model Attributes

## 4. Results

*4.1 Model Evaluation*

To improve the accuracy of the base model, we tried a few transfer-learning models like MobileNet, that gave a validation accuracy of 91.6% and test accuracy of 93.4%. For VGGNet, we have a validation accuracy of 94.1% and test accuracy of 91.4%. For EfficientNet, we were able to achieve a validation accuracy of 84.3% and test accuracy of 80.1%. For ResNet, we were able to achieve a validation accuracy of 67.8% and test accuracy of 70.1% and for InceptionNet, we were able to achieve a validation accuracy of 91.4% and test accuracy of 92.1%. Out of the above tried architectures, Inception Net, VGG and MobileNet gave the best test results. We have chosen MobileNet as our model as it gave the **highest test result** and consumes **smaller model size**.

| Model | Validation Accuracy | Test Accuracy |
|---|---|---|
| MobileNet | 91.60% | 93.40% |
| VGGNet | 94.10% | 91.40% |
| EfficientNet | 84.30% | 80.10% |
| ResNet | 67.80% | 70.10% |
| InceptionNet | 91.40% | 92.10% |

Table 3 Validation accuracy with different models

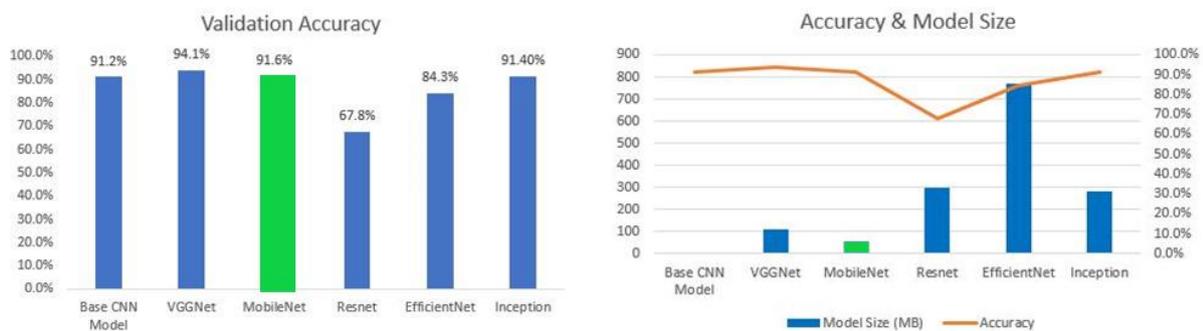

Figure 5 Comparison of various model performances

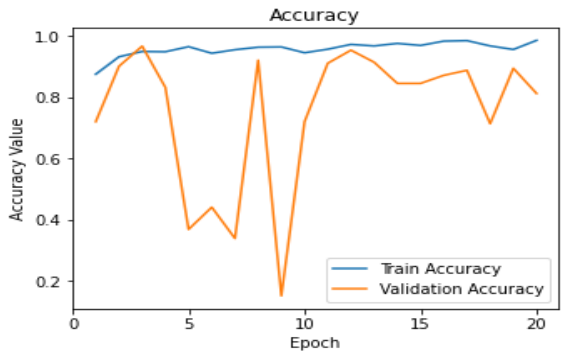

Figure 7 Result graphs of MobileNet model performance

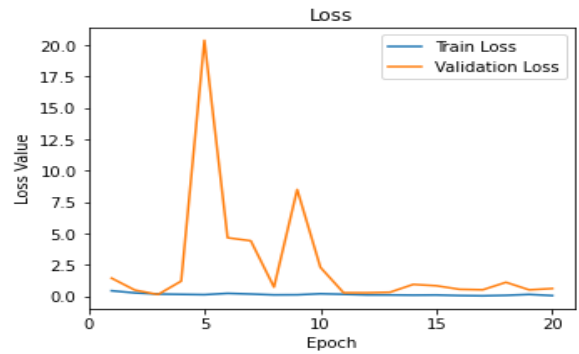

Figure 6 Result graphs of MobileNet model performance

*4.2 Performance results*

As explained in the previous section, our model is based on the MobileNet architecture and have covered the following three use cases successfully

  i) Identify Banana from a given image
  ii) Determine sub-family or variety of Banana and
  iii) Determine the quality of Banana.

For the first and second use case of classifying a given image, the Base CNN model gave a validation accuracy level of 91.2%. Among all the transfer-learning models evaluated, best validation accuracy was attained using our model is at 91.6%. Test accuracy using our model stands at 93.4%. Table 3 is a comparison of model-size and accuracy levels attained using various models. The classification report Table 4 and confusion matrix Table 5 are shown below.

|  | Elakki | Hill Banana | Nendram | Other Fruits | Red Banana | Robusta |
|---|---|---|---|---|---|---|
| precision | 83% | 44% | 100% | 100% | 98% | 96% |
| Recall | 67% | 80% | 77% | 100% | 98% | 100% |
| f1-score | 74% | 57% | 87% | 100% | 98% | 98% |
| support | 15 | 5 | 13 | 22 | 47 | 49 |

Table 4 Classification report for sub-variety use case

|  | **Elakki** | **Hill Banana** | **Nendram** | **Other Fruits** | **Red Banana** | **Robusta** |
|---|---|---|---|---|---|---|
| **Elakki** | 10 | 4 | 0 | 0 | 1 | 0 |
| **Hill Banana** | 0 | 4 | 0 | 0 | 0 | 1 |
| **Nendram** | 1 | 1 | 10 | 0 | 0 | 1 |
| **Other Fruits** | 0 | 0 | 0 | 22 | 0 | 0 |
| **Red Banana** | 1 | 0 | 0 | 0 | 46 | 0 |
| **Robusta** | 0 | 0 | 0 | 0 | 0 | 49 |

Table 5 Confusion matrix for classification of sub-variety

For the third use case of classification based on fruit quality, we have applied this only to the elakki banana variety and have obtained test accuracy of 100%.

|  | **Good Quality** | **Bad Quality** |
|---|---|---|
| **Good Quality** | 5 | 0 |
| **Bad Quality** | 0 | 10 |

Table 6 Confusion matrix for classification of quality of elakki banana

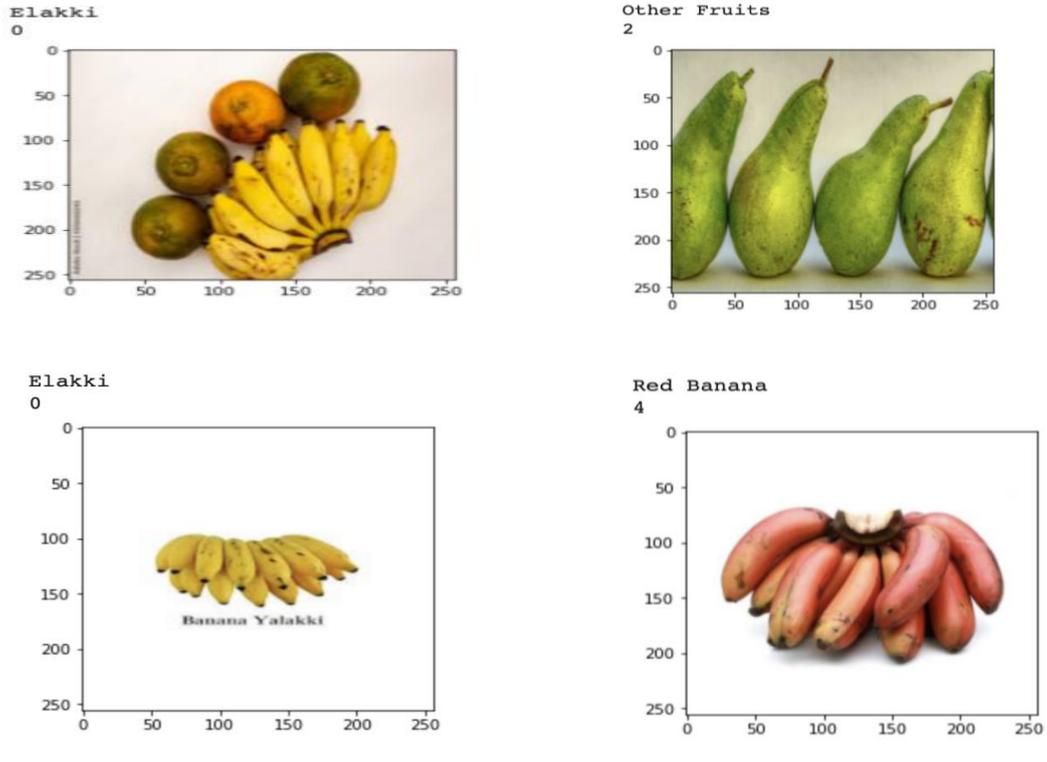

Figure 8 Identify the different varieties of banana

4.3 *Visualizations of results*

For better visualization of the performance of the model we have implemented Grad-CAM technique. Gradient-weighted Class Activation Map (Grad-CAM) for a particular category indicates the discriminative regions used by the CNN to identify that category. This technique makes CNN models more transparent by visualizing the regions of inputs that are most important for predictions from these models. It used the class-specific gradient information flowing into the final convolutional layer of a CNN to produce a coarse localization map of the important regions in the image.

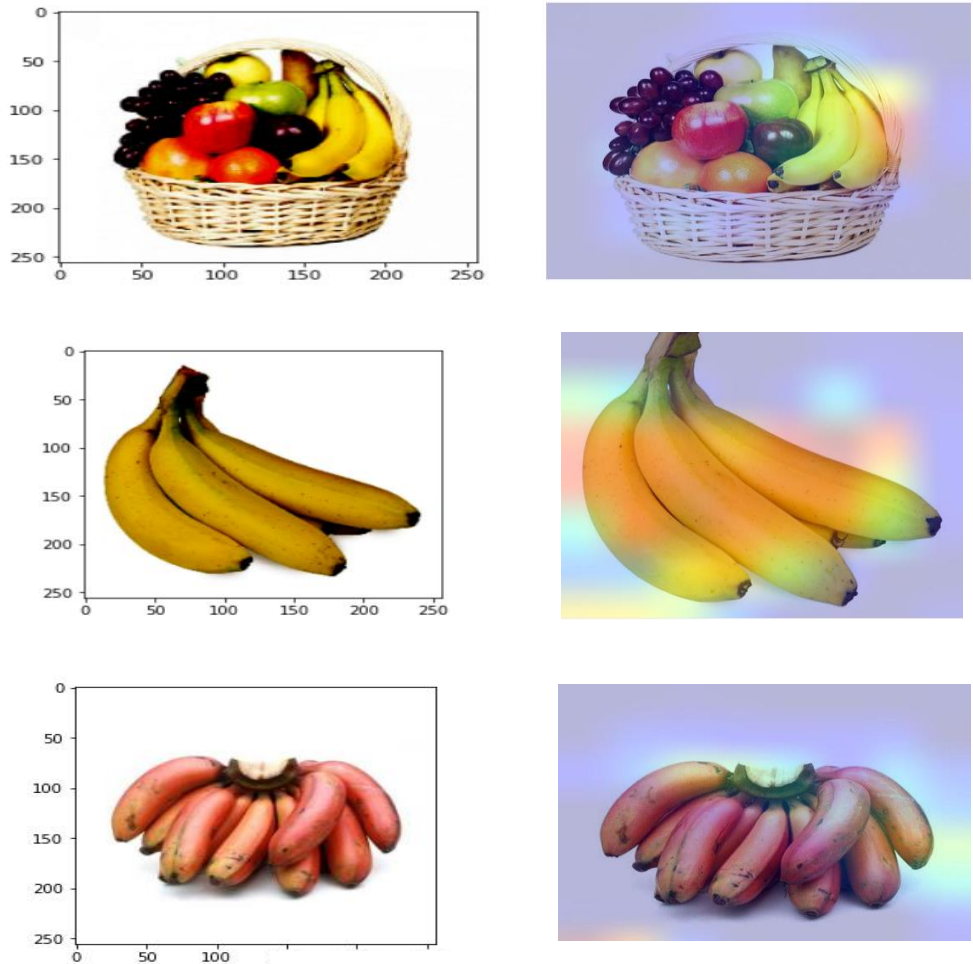

Figure 9 Results of Grad-CAM implementation

**5. Business Implications**

At present, Retail stores rely on cashiers or self-service checkout systems to process the customers' purchases. While most products have barcodes that can be scanned, and hence the checkout time has already been minimized, fruits and vegetables are commonly processed manually. The cashier or the customer need to physically identify and inspect the class of product being bought and look for it in the system before scanning. Visual inspection is labor intensive and prone to human errors and variability.

By adding the ML approach to this process using the model developed by us in this project we can completely automate the process without the need of human intervention for any of the sub-tasks involved.

A short flow of the solution would be as follows: Place one product at a time from basket onto the conveyor belt → click a picture of the item and weigh the product simultaneously → backend model identifies the fruit → assign weight and corresponding price of the item and send to invoice/billing page → Pack and checkout.

We have used raw images (with a white background) without any pre-processing techniques and thereby saving few more seconds of response time. In case of retail stores, the picture clicked would have a black (or any other color of conveyor belt) and can be used as is. A basic camera would suffice as for the hardware requirements. This greatly reduces labor costs and billing counter checkout time while also providing a good customer experience.

We have also included the use case of the model identifying the quality of bananas – good/bad. This would help the procurement and quality check team to classify and remove products that are over-ripe or not of optimum quality. This use-case is a value add to help decrease production costs and increase the quality of the product.

External defects such as surface and skin defect are one of the most influential factors in the commercial quality of fruit, and this is the issue we addressed in our work.

Another feature is identifying and classifying bananas when present with a bunch of other fruits. This would help if the consumer by chance places more than one product/ if two products are placed very close to each other on the belt while the picture is clicked in the billing process.

### 6. Limitations and Future Work

A problem, directly related to the purchase of fruits in retail stores, is that fruits can be inside a plastic bag/ paper bag or any other form of wrap/container. While we have addressed the issue of identifying a single banana/bunch of bananas, we are yet to include the option of capturing these images if present inside any other form of wrapping. Since the regular practice is of first billing followed by packing into bags, we have not explored this option and could be explored as part of future work to make this more practical to use in the market. An addition to further work would be to include quality check for varieties other than Elakki and to test for two or more varieties of bananas in the same frame. Another drawback to be considered is that our model would classify fake bananas or any object with the shape and color of a banana as a real fruit. This is due to the fact that we have assumed only real fruit images would be fed into the model and have limited its working to identifying skin/surface level features/defects only. Further on the model deployment and production, creating a REST api, predicting and maintenance are part of future work of this paper.

### 7. Conclusion

For this project we have explored various options to come up with a solution using computer vision. With the help of transfer learning using other advanced architectures such as VGG Net, ResNet, Inception Net, EfficientNet and Mobile Net on top of a base CNN model and after rigorous testing we converged on classification using MobileNet as the best option. There are three main points we would like to discuss as a closing note of this paper,
1) Hardware requirements - As for the hardware required for deployment in businesses, the model can be deployed on commodity hardware and is suitable for mid-size retail stores as they can leverage this product without huge setup costs.
2) Part of overall inventory management - The model can be used as part of overall inventory automation. First by the procurement team for quality pass and also it could be integrated with billing counter system to auto detect the type of banana and then calculate the price as per the weight.
3) Scalability and Future Vision - Overall the model can be scaled and trained to predict and classify other varieties of fruits, vegetables and other objects in retail outlets.